\documentclass{article}

    \PassOptionsToPackage{numbers, compress}{natbib}

\usepackage[preprint]{neurips_2024}

\usepackage[utf8]{inputenc} 
\usepackage[T1]{fontenc}    
\usepackage{hyperref}       
\usepackage{url}            
\usepackage{booktabs}       
\usepackage{amsfonts}       
\usepackage{nicefrac}       
\usepackage{microtype}      
\usepackage{xcolor}         
\usepackage{amsmath}
\usepackage{mathtools}

\usepackage{multirow}

\usepackage{caption}
\usepackage{subcaption}

\usepackage{color, soul}

\usepackage{enumitem}
\usepackage{algorithm}
\usepackage{algpseudocode}

\title{Overcoming Catastrophic Forgetting in\\
Federated Class-Incremental Learning via\\
Federated Global Twin Generator}

\author{%
  Thinh Nguyen$^1$, Khoa D Doan$^1$, Binh T. Nguyen$^2$, Danh Le-Phuoc$^3$, Kok-Seng Wong$^1$ \\
  $^1$College of Engineering \& Computer Science, VinUniversity, Hanoi, Vietnam\\
  $^2$Vietnam National University - Ho Chi Minh city, University of Science, Vietnam\\
  $^3$Technische Universität Berlin, Germany\\
  \texttt{\{thinh.nth, khoa.dd, wong.ks\}@vinuni.edu.vn} \\
  \texttt{ngtbinh@hcmus.edu.vn}, \texttt{danh.lephuoc@tu-berlin.de}
}

\begin{document}

\maketitle

\begin{abstract}
  Federated Class-Incremental Learning (FCIL) increasingly becomes important in the decentralized setting, where it enables multiple participants to collaboratively train a global model to perform well on a sequence of tasks without sharing their private data. In FCIL, conventional Federated Learning algorithms such as FedAVG often suffer from catastrophic forgetting, resulting in significant performance declines on earlier tasks. Recent works, based on generative models, produce synthetic images to help mitigate this issue across all classes, but these approaches' testing accuracy on previous classes is still much lower than recent classes, i.e., having better plasticity than stability. To overcome these issues, this paper presents Federated Global Twin Generator (FedGTG), an FCIL framework that exploits privacy-preserving generative-model training on the global side without accessing client data. Specifically, the server trains a data generator and a feature generator to create two types of information from all seen classes, and then it sends the synthetic data to the client side. The clients then use feature-direction-controlling losses to make the local models retain knowledge and learn new tasks well. We extensively analyze the robustness of FedGTG on natural images, as well as its ability to converge to flat local minima and achieve better-predicting confidence (calibration). Experimental results on CIFAR-10, CIFAR-100, and tiny-ImageNet demonstrate the improvements in accuracy and forgetting measures of FedGTG compared to previous frameworks.
\end{abstract}

\section{Introduction}
\label{sec:intro}

\begin{figure}[t]
     \centering
     \includegraphics[scale=0.6]{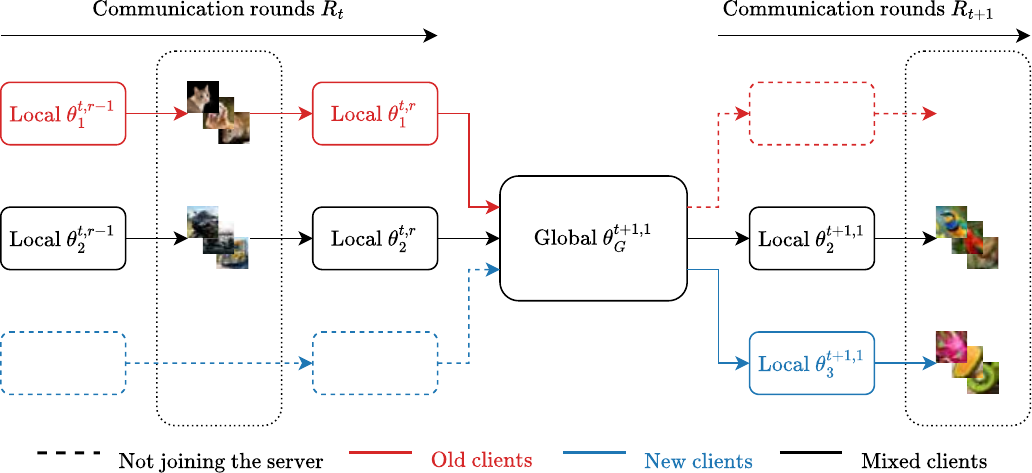}
     \vspace{0.2cm}
    
    \caption{Illustration of the real-world scenarios in the FCIL setting.}
    \label{fig:fcil}
\end{figure}

Federated Learning (FL)~\cite{mcmahan2016federated, bonawitz2019towards} is a Machine Learning setting that facilitates collaborative learning while maintaining privacy. Despite its significant achievements on various domains~\cite{doshi2022federated, lin2021fednlp, liu2021federated, nguyen2019diot}, FL observes several critical challenges, including resource limitation and data heterogeneity. Moreover, the client's local data distribution is assumed to remain unchanged, but the real-world scenarios~\cite{aljundi2019continual} can be totally different, where the client's task, data, and domain can be changed, as shown in Figure~\ref{fig:fcil}. To overcome such challenges, Federated Class-Incremental Learning (FCIL)~\cite{dong2022federated, dong2023no} is an innovative approach that combines the principles of FL and Class-Incremental Learning (CIL)~\cite{rebuffi2017icarl} to enable models to learn continuously from decentralized data sources while adapting to new information over time without forgetting previous knowledge~~\cite{french1999catastrophic}. This approach addresses data privacy challenges and ensures the model can evolve as new data types or tasks emerge without needing to access historical data. In CIL, exemplar-based approaches~\cite{rebuffi2017icarl, chaudhry2019tiny, buzzega2020dark} preserve a limited number of samples from previous tasks to prevent forgetting, whereas exemplar-free approaches~\cite{he2018exemplar, liu2020generative, magistri2024elastic} do not retain any samples from prior tasks.

\begin{figure*}[t]
    \centering
    \includegraphics[scale=0.65]{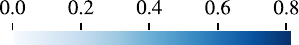}\\
    \vspace{0.3cm}
     \begin{subfigure}[b]{0.22\textwidth}
         \centering
         \includegraphics[scale=0.5]{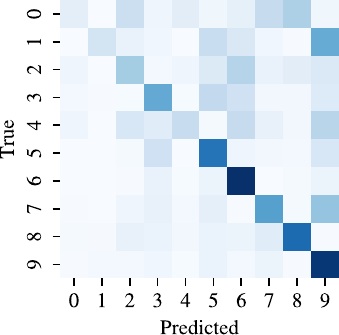}\\
         \caption{TARGET}
         \label{fig:heatmap:target}
     \end{subfigure}
     \hfill
     \begin{subfigure}[b]{0.22\textwidth}
         \centering
         \includegraphics[scale=0.5]{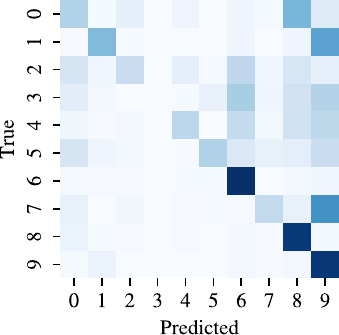}\\
         \caption{MFCL}
         \label{fig:heatmap:mfcl}
     \end{subfigure}
     \hfill
     \begin{subfigure}[b]{0.22\textwidth}
         \centering
         \includegraphics[scale=0.5]{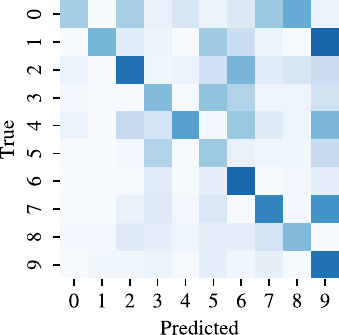}\\
         \caption{$G_D+G_F$}
         \label{fig:heatmap:gdgf}
     \end{subfigure}
     \hfill
     \begin{subfigure}[b]{0.22\textwidth}
         \centering
         \includegraphics[scale=0.5]{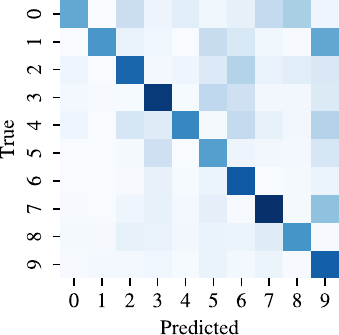}\\
         \caption{FedGTG (Our)}
         \label{fig:heatmap:fedgtg}
     \end{subfigure}
        \caption{Confusion matrix among FCIL algorithms: \textbf{(a)} TARGET, \textbf{(b)} MFCL, \textbf{(c)} only the application of two generators to FL, and \textbf{(d)} FedGTG, testing on CIFAR-10 after training is completed. While TARGET and MFCL have bad predicting performance on initial classes and two generators struggle to learn new tasks, FedGTG achieves a better stability-plasticity trade-off.} 
        \label{fig:heatmap}
\end{figure*}

Due to privacy concerns in many FL systems (e.g., hospitals and research areas), users cannot store any historical data, so the exemplar-free category is of particular interest. Recent works~\cite{zhang2023target, qi2023better, babakniya2024data} in this field tend to generate synthetic data and combine with distilling regularizers~\cite{hinton2015distilling, liu2020generative} to balance the trade-off between retaining knowledge and learning new tasks. However, experimental results have shown that these works still witness catastrophic forgetting, i.e., bias towards recent classes; see Figure~\ref{fig:heatmap:target} and~\ref{fig:heatmap:mfcl}.


To address this problem, we propose \textbf{Federated Global Twin Generator} (FedGTG), an FCIL framework that does not store any data from clients. After completing one task,  the server trains two generative models and shares them with clients on subsequent tasks to create synthetic examples and features of previous classes. On the client side, we propose a synthetic logit distillation using generated features for retaining knowledge, as well as a fine-tuning loss using both real and generated data to balance the ability to predict all classes. However, using only these two objectives still hinders the model's ability to obtain new knowledge, see Figure~\ref{fig:heatmap:gdgf}. We argue that this issue happened as the feature directions were not constrained. Therefore, we add a feature-direction-controlling loss, which helps the model to have more plasticity in learning new tasks. As a result, FedGTG achieved extensive performance in terms of accuracy and forgetting measures compared to previous methods, as shown in Section~\ref{sec:result}.

We summarize our contributions below:

\begin{itemize}[leftmargin=*]
\item We propose an FCIL framework ensuring clients' privacy by training a data generator and a feature generator on the server side. These generators are distributed to the clients to mitigate catastrophic forgetting. 

\item On the client side, we propose direction-controlling objectives to help the model have a better stability-plasticity trade-off.


\item We demonstrate the effectiveness of our method in popular benchmarks, including CIFAR-10, CIFAR-100, and tiny-ImageNet, with a 4\% increase in accuracy and a 10\% decrease in forgetting.

\item We analyze the robustness of FedGTG compared with recent FCIL algorithms on natural images, as well as test its abilities to converge to flat minima and achieve better predicting confidence.


\end{itemize}

\section{Related work and preliminaries}

\subsection{Continual Learning}
Catastrophic forgetting~\cite{french1999catastrophic} is a major issue in machine learning where training a model on new data makes it forget its previously learned knowledge. This issue is central to the field of CL, whose primary objective is to build models to acquire new knowledge while retaining information from older tasks. Numerous strategies have been explored to mitigate this problem, including the implementation of regularization terms~\cite{li2017learning, kirkpatrick2017overcoming, zenke2017continual}, the isolation of architectural parameters~\cite{mallya2018packnet, yoon2017lifelong, mallya2018piggyback},  the use of experience replay through data storage~\cite{rebuffi2017icarl, chaudhry2019tiny, buzzega2020dark}, and studies of data generation~\cite{he2018exemplar, liu2020generative, magistri2024elastic}. In CL, replay-based methods observe significant performance in terms of accuracy and forgetting measures. However, due to privacy concerns in FCIL,  privacy concerns prevent data storage, making these methods inapplicable. An extensive alternative to address this issue is generative-based approaches.

\paragraph{Learning tasks} There are three main types of learning in CL: Task-Incremental Learning (Task-IL), Domain-Incremental Learning (Domain-IL), and Class-Incremental Learning (Class-IL)~\cite{van2019three}. In Task-IL, each task is distinct and comes with its own distribution during training and testing. In Domain-IL, the learning task does not change, while different domains or distributions of data sequentially arrive during training. In Class-IL, each new task adds new classes to what the model has to learn, which continually expands the amount of information the model needs to handle.

\subsection{Federated Class-Incremental Learning}

The goal of FCIL is to train a model to learn new classes over time without forgetting previously learned classes while also ensuring that data privacy is maintained across multiple decentralized devices. Several approaches exploit Knowledge Distillation~\cite{hinton2015distilling} to mitigate forgetting by appointing the global model's weight of the most recent task as a teacher. \textbf{Continual Federated Learning with Distillation (CFeD)}~\cite{ma2022continual} constructs server and client-side knowledge distillation using a surrogate dataset, but this costs time and financial resources to collect enough data for this surrogate dataset. \textbf{Global-Local Forgetting Compensation (GLFC)}~\cite{dong2022federated} relaxes this problem by training a proxy server globally to ease the imbalance issue between classes.

The above approaches yield extensive performance in past knowledge retention but lack the ability to learn well on new tasks. To alleviate this issue, \textbf{Federated Class-Incremental Learning (FedCIL)}~\cite{qi2023better} trains generators at both client-side and server-side, as well as utilizing knowledge distillation to balance the stability-plasticity trade-off. However, this approach can raise a privacy risk since information about client-side generative models is shared with the server. \textbf{Federated Class-Continual Learning via Exemplar-Free Distillation (TARGET)}~\cite{zhang2023target} and \textbf{Mimicking Federated Continual Learning (MFCL)}~\cite{babakniya2024data} ensures clients' privacy and limited storage by training a data generator on the global-side and adding distilling regularizers to the client-side training to enhance overall performance. However, as we show in Figure~\ref{fig:heatmap}, these methods still have bad performance on old classes, leaving catastrophic forgetting mitigation a still desirable goal.

\subsection{Preliminaries}

There are $c$ clients, denoted as $\left\{C_1, C_2, \ldots, C_c\right\}$ and a central server, denoted as $S$. We consider the Synchronous Federated Continual Learning setting \cite{yang2024federated} where all clients share the same task sequence $T=\left\{T_1, T_2, \ldots, T_n\right\}$. Specifically, at task $T_t$, each client $C_i$ has a private dataset $\mathcal{D}^t_i=\left\{\mathcal{X}^t_i, \mathcal{Y}^t\right\}$.

During the first task, the global model $\theta^1_G$ is obtained after aggregating local models $\left\{\theta^1_1, \theta^1_2, \ldots, \theta^1_{s_1}\right\}$ using conventional Deep Learning methods, where $s_1$ is a number of selected clients among all. From the task $T_t$, $t\geq 2$, the global model $\theta^{t-1}_G$ can distinguish the samples belonged to the classes set $\bigcup_{i=1}^{t-1}\mathcal{Y}_i$. The server then distributes its parameters back to the clients. Client $C_i$ uses $\theta^{t-1}_G$ as an initial model to train on task $T_t$ using its private dataset $\mathcal{D}^t_i$. The local model $\theta^t_i$ should perform well in classifying classes from the set $\bigcup_{i=1}^{t}\mathcal{Y}_i$. Finally, the server collects the local models from clients who participate in the process after each $r_t$ communication round and obtains a new global model $\theta^t_G$, which can identify classes from the set $\bigcup_{i=1}^{t}\mathcal{Y}_i$.

\section{Metholodgy}



\subsection{Overview}
\label{sec:overview}
\begin{figure}[t]
  \centering
  \includegraphics[scale=0.53]{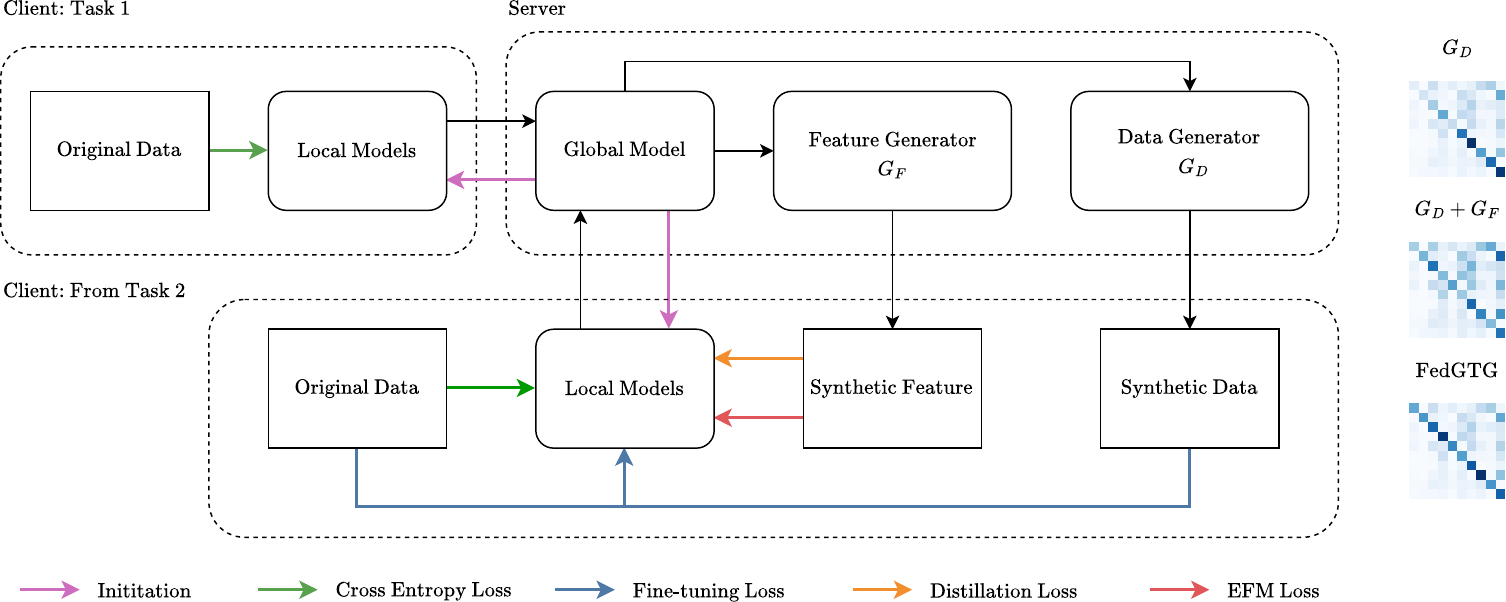}
  \caption{Illustration of the proposed framework. After completing one task, the server employs a data-free approach to train two generators. The clients then use two types of synthetic information from these generators to train their local models for retaining knowledge and learning new tasks well.}
  \label{fig:fedgtg}
\end{figure}

Several approaches~\cite{rebuffi2017icarl, chaudhry2019tiny, buzzega2020dark} in the conventional CL achieve significant predicting performance across all classes by storing a small subset of samples from previous tasks in episodic memory. However, this approach is not viable in the FL setting due to privacy concerns (e.g., local hospitals cannot share data with the central server). One initial approach to address this challenge is using generative models, which can generate synthetic data for subsequent training, as demonstrated in earlier studies \cite{zhang2023target, babakniya2024data}. However, as illustrated in Figure~\ref{fig:heatmap}, only generating synthetic examples still causes catastrophic forgetting in previously learned classes. Therefore, we propose a Federated Global Twin Generator (FedGTG), which can balance the stability-plasticity trade-off and clients' privacy. This method has two main stages: (1) At the end of each task, the server trains a data generator and a feature generator to capture the information of all seen classes; (2) Clients receive generators from the server to create synthetic information, as well as obtaining global weights as initialization, which helps retain knowledge from previous tasks and learn the new task efficiently. In the following section, we will explain these two stages of FedGTG (Figure \ref{fig:fedgtg}).

\subsection{Server-side}


\subsubsection{Data Generator}

Since the server only has access to the global model's weights, we can only train the data generator using data-free methods, such as DeepInversion. Specifically, we have a generative model that takes a noise $z\sim\mathcal{N}\left(0, 1\right)$ as input and produces a synthetic example $\tilde{x}$ mirroring the dimensions of the original training input. This synthetic data should observe the following objectives.

After training task $T_t$, the synthetic data should be classified correctly by the global model $\theta^t_G$, and they should not be biased to any classes. With this aim, we employ a temperature cross entropy classification loss between its assigned label $z$ and the prediction of $\theta^t_G$ on $G^t_D\left(z\right)$ as
\begin{equation}
\mathcal{L}_{\text{CE}} = \text{CE}_{\text{last}}\left(\text{argmax}\left(z\left[:, q\right]\right), \theta^t_G\left(\widetilde{x}\right)\right) + \lambda_{\text{current}}\text{CE}_{\text{current}}\left(\text{argmax}\left(z\left[:, q\right]\right), \theta^t_G\left(\widetilde{x}\right)\right),
\end{equation}
where $\widetilde{x}$ is the output of $G^t_D\left(z\right)$, $\text{CE}_{\text{last}}$ and $\text{CE}_{\text{current}}$ respectively are the Cross-Entropy Loss using the truncated outputs of $\theta^t_G\left(\widetilde{x}\right)$ corresponding with last classes from task $T_i$, $i<t$, and current learned classes on task $T_t$, and $\lambda_{\text{current}}$ is the temperature hyper-parameter.

Generating synthetic examples can easily be biased to a subset of classes. To maintain the diversity between classes, we utilize the Information Entropy (IE) Loss~\cite{chen2019data} as follows:
\begin{equation}
\mathcal{L}_{\text{IE}}=-\mathbf{H}_{\text{info}}\left(\frac{1}{\text{bs}}\sum_{i=1}^{\text{bs}}\theta^t_G\left(\widetilde{x}_i\right)\right), \text{bs: batch size}, 
\label{loss:ie}
\end{equation}
This loss measures the IE for samples of a batch. By maximizing this value, we can promote a more uniform and balanced output distribution from the generator across all classes.

In order to further improve the stability of generator training, we use Batch Normalization Loss~\cite{smith2021always} to  make all Batch Normalization layers have the same statistics on synthetic images, as follows:

\begin{equation}
\mathcal{L}_{\text{batch}}=\frac{1}{L}\sum_{j=1}^L \mathbf{KL}\left(\mathcal{N}\left(\mu_j, \sigma_j^2\right) \parallel \mathcal{N}\left(\widetilde{\mu}_j, \widetilde{\sigma}^2_j\right)\right),
\end{equation}
where $L$ is the total number of Batch Normalization layers in the architecture of the global model. $\mu_j$ and $\sigma_j^2$ are the mean and variance stored in Batch Normalization layer $j$ of the global model, $\widetilde{\mu}_j$ and $\widetilde{\sigma}_j$ are measured statistics of Batch Normalization layer $j$ for the synthetic data.
 
Adjacent pixels in real images typically have values that are near to one another.  One typical method to promote similar patterns in the synthetic images is to add Image Prior Loss~\cite{haroush2020knowledge}. We can create the smoothed (blurred) version of an image by applying a Gaussian kernel and minimizing the distance of the original and $\text{Smooth}\left(\widetilde{x}\right)$ as
\begin{equation}
\mathcal{L}_{\text{smooth}}=\left\lVert \widetilde{x} - \text{Smooth}\left(\widetilde{x}\right) \right\rVert^2_2.
\end{equation}

In summary, we can write the training objective of $G_D$ as follows:
\begin{equation}
\min_{G_D} \mathcal{L}_{\text{CE}} + \lambda_{\text{IE}}\mathcal{L}_{\text{IE}}+\lambda_{\text{batch}}\mathcal{L}_{\text{batch}}+\lambda_{\text{smooth}}\mathcal{L}_{\text{smooth}},
\end{equation}
where $\lambda_{\text{IE}}$, $\lambda_{\text{batch}}$, and $\lambda_{\text{smooth}}$ are hyper-parameters of specific loss functions.

\subsubsection{Feature Generator}

As mentioned in Section~\ref{sec:overview}, only synthetic images can exacerbate the catastrophic forgetting problem. To address this, we train a feature generator that synthesizes features, capturing the knowledge within the feature space. Like the data generator, this generative model is trained only on the server side. The feature generator takes noise input $z\sim\mathcal{N}\left(0,1\right)$ and produces synthetic features $\tilde{f}$ that match the dimensions of the original features. These synthetic features must meet the following objectives:

\paragraph{Temperatured Cross Entropy Loss}
After training task $T_t$, the generative feature should be classified correctly by the classifier $H^t_G$ of the global model. Additionally, the synthetic features should not be biased to any classes. With this aim, we employ a temperature cross entropy classification loss between its assigned label $z$ and the prediction of $H^t_G$ on $G^t_D\left(z\right)$ as
\begin{equation}
\mathcal{L}_{\text{FCE}} = \text{CE}_{\text{last}}\left(\text{argmax}\left(z\left[:, q\right]\right), H^t_G\left(\widetilde{f}\right)\right) + \lambda_{\text{current}}\text{CE}_{\text{current}}\left(\text{argmax}\left(z\left[:, q\right]\right), H^t_G\left(\widetilde{f}\right)\right),
\end{equation}
where $\widetilde{f}$ is the output of $G^t_F\left(z\right)$, $\text{CE}_{\text{last}}$ and $\text{CE}_{\text{current}}$ respectively are the Cross Entropy Loss using the truncated outputs of $H^t_G\left(\widetilde{f}\right)$ corresponding with last classes from task $T_i$, $i<t$, and current learned classes on task $T_t$, and $\lambda_{\text{current}}$ is the temperature hyper-parameter.

\paragraph{Feature Information Entropy Loss} The generated features should also not be biased to any subset of classes. Therefore, we adjust the IE Loss (Equation~\ref{loss:ie}) and propose the Feature Information Entropy Loss to make the synthetic feature have this quality, which is
\begin{equation}
\mathcal{L}_{\text{FIE}}=-\mathbf{H}_{\text{info}}\left(\frac{1}{\text{bs}}\sum_{i=1}^{\text{bs}}H^t_G\left(\widetilde{f}_i\right)\right), \text{bs: batch size},
\end{equation}
In summary, we can write the training objective of $G_F$ as follows:
\begin{equation}
\min_{G_F} \mathcal{L}_{\text{FCE}} + \lambda_{\text{FIE}}\mathcal{L}_{\text{FIE}},
\end{equation}
where $\lambda_{\text{FIE}}$ is the hyper-parameter of Feature Information Entropy Loss.

\subsection{Client-side}

From task $T_{t\geq 2}$, on the client side, we have to solve the trade-off between learning the current task quickly (plasticity) and retaining knowledge from previous tasks efficiently (stability). To this end, we divide the learning objectives into two parts.

\paragraph{Plasticity}
To learn new tasks well, the model needs to learn the new information in a way that is separate from the old classes. To do this, we compute the Cross-Entropy Loss by using only the new classes' linear heads. Formally, we minimize:

\begin{equation}
\mathcal{L}_{\text{CE}}=
\text{CE}\left(\theta^t\left(x\mid T_t\right), y\right),
\label{loss:ce}
\end{equation}
where $\theta^t\left(x\mid T_t\right)$ is model's output and masking old classes before task $T_t$'s linear heads.

\paragraph{Stability} To mitigate forgetting, previous approaches leverage knowledge distillation~\cite{usmanova2021distillation, zhang2023target}. However, this can cause information loss in probability space due to squashing functions~\cite{liu2018improving}. Therefore, motivated by Buzzega et al.~\cite{buzzega2020dark}, we propose Synthetic Logits Distillation Loss, which matches the logits of the old and current linear heads. These classifiers take synthetic features as input instead of synthetic data since the feature stores more previous information. Formally, we optimize:
\begin{equation}
\mathcal{L}_{\text{logits}} = \left\lVert \theta^{t-1}_G\left(\widetilde{f}\right) - \theta^{t}\left(\widetilde{f}\right)\right\rVert,
\label{loss:logit}
\end{equation}
where $\theta^{t-1}_G$ is the global model trained up to task $T_{t-1}$.

As shown in~\cite{buzzega2020dark}, when there is a sudden shift in the distribution of the input of the task sequence, biased features on previous tasks can output biased logits, hindering the ability to obtain new knowledge. To mitigate this shortcoming, from task $T_t$, we leverage synthetic data by training it alongside with real data. However, the distribution of synthetic data may vary from that of real data, making it essential to ensure the model does not distinguish between old and new data solely based on these differences. To address this problem, we only use the extracted features of the data. To tackle this issue, we utilize only the extracted features of the data, i.e., clients freeze the feature extraction layers and update only the linear head (represented by $H^t$) for both real ($x$) and synthetic ($\widetilde{x}$) images. This Fine-tuning loss is formulated as
\begin{equation}
\mathcal{L}_{\text{FT}}=\text{CE}\left(H^t\left(\left[f, \widetilde{f}\right]\right), \left[y, \widetilde{y}\right]\right),
\label{loss:ft}
\end{equation}
where $f$ and $\widetilde{f}$ respectively are the extracted features of $x$ and $\widetilde{x}$ after passing through the freezed feature extractor $F^t$ of the local model, $y$ and $\widetilde{y}$ is the hard label of $x$ and $\widetilde{x}$.

\paragraph{Enhancing stability-plasticity}
Figure~\ref{fig:heatmap:gdgf} shows that the combination of the above objectives reduces the model's performance across all classes. We contend that this happens because the feature directions are unconstrained, resulting in the total loss failing to converge. We then further add additional loss to balance this problem, named Empirical Feature Matrix Loss \cite{magistri2024elastic}, which constrains directions in feature space most important for previous tasks, while it allows more plasticity in other directions when learning new tasks. In this work, we re-utilize the synthetic features to calculate the Empirical Feature Matrix $E_{t-1}$ from the previous task $T_{t-1}$. We have,
\begin{equation}
\mathcal{L}_{\text{EFM}}=\left(F^t\left(x\right)-F^{t-1}_G\left(x\right)\right)^\top \left(\lambda_{E}E_{t-1}+\eta I\right) \left(F^t\left(x\right)-F^{t-1}_G\left(x\right)\right),
\label{loss:efm}
\end{equation}
where $F^t$ and $F^{t-1}_G$ respectively are the feature extractor of the current model and the previous global model, $\eta$ is the damping term to constrain features to stay in a specific region.

In summary, the final objective on the client side as
\begin{equation}
\min_{\theta^t}\mathcal{L}_{\text{CE}}+\lambda_{\text{logits}}\mathcal{L}_{\text{logits}}+\lambda_{\text{FT}}\mathcal{L}_{\text{FT}}+\lambda_{\text{EFM}}\mathcal{L}_{\text{EFM}}.
\label{loss:optimize}
\end{equation}

\section{Experimental results}
\label{sec:exp_result}
\subsection{Experimental Setup}
\label{sec:implementation}

In this section, we provide our experimental setup, including the datasets used, FCIL baselines, and implementation details.

\paragraph{Datasets} We perform our experiments on three widely-used benchmark datasets in FCIL~\cite{dong2022federated, zhang2023target, babakniya2024data}, which are the protocol versions in the FCIL setting of \textbf{CIFAR-10}~\cite{krizhevsky2009learning}, \textbf{CIFAR-100}~\cite{krizhevsky2009learning}, \textbf{tiny-ImageNet}~\cite{yao2015tiny}, and we name it respectively are \textbf{Sequential F-CIFAR-10}, \textbf{Sequential F-CIFAR-100} and \textbf{Sequential F-tiny-ImageNet}. The data preparation is explained later in Appendix~\ref{ap:setup}. We use Latent Dirichlet Allocation (LDA)~\cite{reddi2020adaptive} with $\alpha=1$ to distribute the data of each task among clients.



\paragraph{FCIL Baselines} In addition to our FedGTG, we also include one regularization-based method, {\bf FLwF-2T}~\cite{usmanova2021distillation}, and two generative-based methods, {\bf TARGET}~\cite{zhang2023target} and {\bf MFCL}~\cite{babakniya2024data}. The detailed description of these algorithms can be seen in Appendix~\ref{ap:setup}.

\paragraph{Models and Implementation Details} In all experiments, we train a ResNet-18 \cite{he2016deep} backbone using the SGD optimizer~\cite{bottou1998online}. We train the model for $100$ epochs per task on every dataset. Additional implementation details and hyper-parameter configurations are then provided in the Appendix~\ref{ap:setup}.




\paragraph{Evaluation Metrics} We report the performance of the methods using two metrics: Average Incremental Accuracy and Average Forgetting. {\bf Average Incremental Accuracy (AIA)} measures the average accuracy of the global model on all tasks after the training finishes. Forgetting ($f_t$)  of task $T_t$ is the difference between the model's best performance on task $T_t$ and its accuracy after completed training. Consequently, {\bf Average Forgetting (AF)} is the average of all $f^t$, from task $T_1$ to task $T_{n-1}$, at the end of task $T_n$. We report the averaged result over three different random initializations.

\subsection{Performance Results}
\label{sec:result}

\begin{figure*}[t]
    \centering
    \includegraphics[scale=0.65]{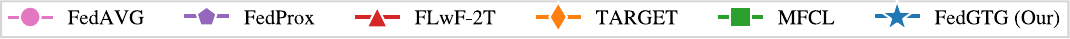}\\
    \vspace{0.2cm}
     \begin{subfigure}[b]{0.30\textwidth}
         \centering
         \includegraphics[scale=0.45]{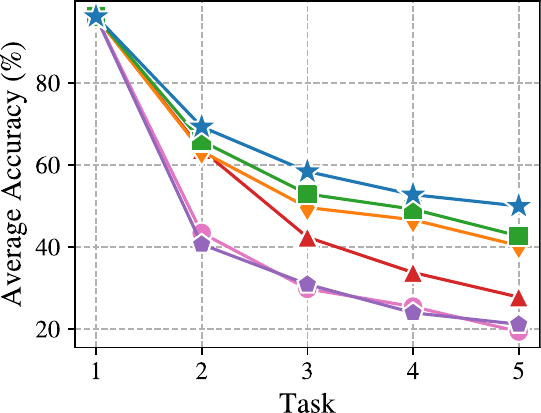}\\
         \caption{Sequential F-CIFAR-10}
         \label{fig:acc:cifar10}
     \end{subfigure}
     \hfill
     \begin{subfigure}[b]{0.30\textwidth}
         \centering
         \includegraphics[scale=0.45]{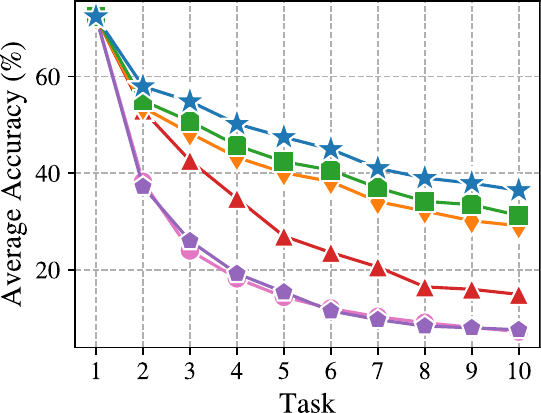}
         \caption{Sequential F-CIFAR-100}
         \label{fig:acc:cifar100}
     \end{subfigure}
     \hfill
     \begin{subfigure}[b]{0.30\textwidth}
         \centering
         \includegraphics[scale=0.45]{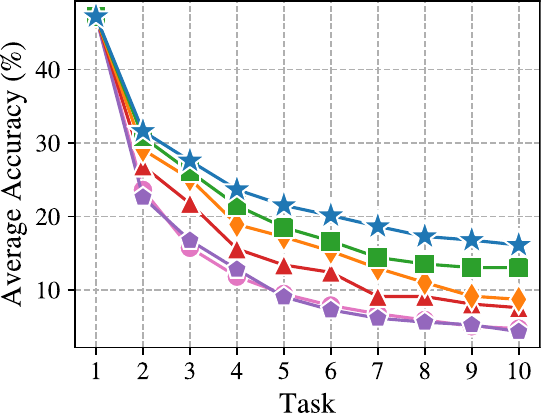}
         \caption{Sequential F-tiny-ImageNet}
         \label{fig:acc:tinyimg}
     \end{subfigure}
        \caption{Average Accuracy per task of various algorithms on popular benchmarks.} 
        \label{fig:acc}
\end{figure*}

\begin{table}[t]
\centering
\caption{Performance of the different baselines in terms of AIA and AF for three datasets. [$\uparrow$] higher is better, [$\downarrow$] lower is better.}
\vspace{0.1cm}
\resizebox{\textwidth}{!}{
\begin{tabular}{ccccccc}
\hline
\multirow{2}{*}{\textbf{Method}} & \multicolumn{2}{c}{\textbf{CIFAR-10}}                & \multicolumn{2}{c}{\textbf{CIFAR-100}}              & \multicolumn{2}{c}{\textbf{tiny-ImageNet}}          \\ \cline{2-7} 
                                 & AIA [$\uparrow$]         & AF [$\downarrow$]         & AIA [$\uparrow$]         & AF [$\downarrow$]        & AIA [$\uparrow$]         & AF [$\downarrow$]        \\ \hline
FedAVG                           & $42.82\pm 0.23$          & $55.55 \pm 0.58$          & $21.39\pm 0.22$          & $78.67\pm 0.83$          & $13.80\pm 0.19$          & $74.12\pm 0.81$          \\
FedProx                          & $42.43\pm 0.32$          & $56.15\pm 0.71$           & $21.54\pm 0.32$          & $78.12\pm 0.71$          & $13.69\pm 0.21$          & $75.16\pm 0.79$          \\
FLwF-2T                          & $52.74\pm 0.23$          & $39.51\pm 0.59$           & $32.19\pm 0.18$          & $50.78\pm 0.63$          & $17.18\pm 0.17$          & $44.51\pm 0.67$          \\
TARGET                           & $59.19\pm 0.16$          & $17.23 \pm 0.45$          & $42.15\pm 0.13$          & $26.45\pm 0.61$          & $19.46\pm 0.25$          & $20.17\pm 0.57$          \\
MFCL                             & $61.34\pm 0.21$          & $22.32 \pm 0.52$          & $45.07\pm 0.12$          & $28.30\pm 0.78$          & $21.47\pm 0.15$          & $23.90\pm 0.58$          \\
\textbf{FedGTG (Our)}           & $\mathbf{64.50\pm 0.22}$ & $\mathbf{13.14 \pm 0.67}$ & $\mathbf{46.42\pm 0.18}$ & $\mathbf{18.66\pm 0.76}$ & $\mathbf{24.04\pm 0.23}$ & $\mathbf{16.18\pm 0.62}$ \\ \hline
\end{tabular}
}
\label{tab:acc}
\end{table}

We present the performance of FedGTG and the baselines. Figure~\ref{fig:acc} shows the Average Accuracy of the model at each task in the training process. It can be seen that FedGTG achieves state-of-the-art performance in all settings. Specifically, our method observes better Average Accuracy on all later tasks. Table~\ref{tab:acc} reports both AIA results \textit{(higher is better)} and AF results \textit{(lower is better)}.

As expected, FedAvg and FedProx suffer the highest forgetting since they are not designed for FCIL. When compared to FLwF-2T, the performance gap between it and our FedGTG is significant, indicating that regularization towards previous parameter sets is insufficient to avoid forgetting. Compared to the generative-replay methods, TARGET and MFCL, our FedGTG achieves the least AF and the best AIA, showing that FedGTG can both retain knowledge and learn new tasks effectively.



\subsection{Model Analysis}
\label{sec:analysis}
The majority of FCIL research concentrates on testing experiments on ideal benchmarks~\cite{dong2022federated, zhang2023target, babakniya2024data}, such as CIFAR~\cite{krizhevsky2009learning} and ImageNet~\cite{deng2009imagenet}. This results in a lack of analysis concerning real-world scenarios, such as the decision-making ability required in hospitals or the model's generalization to diverse environments. Therefore, in this section, we conducted experiments to analyze the robustness of FedGTG and three FCIL algorithms mentioned above on corrupted environments, as well as the qualities of generalization~\cite{chaudhari2019entropy, jastrzkebski2017three, keskar2016large} and achieve calibrated networks~\cite{guo2017calibration, kull2019beyond}.

\begin{figure*}[t]
    \centering
    \includegraphics[scale=0.65]{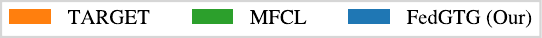}\\
    \vspace{0.2cm}
    \includegraphics[scale=0.45]{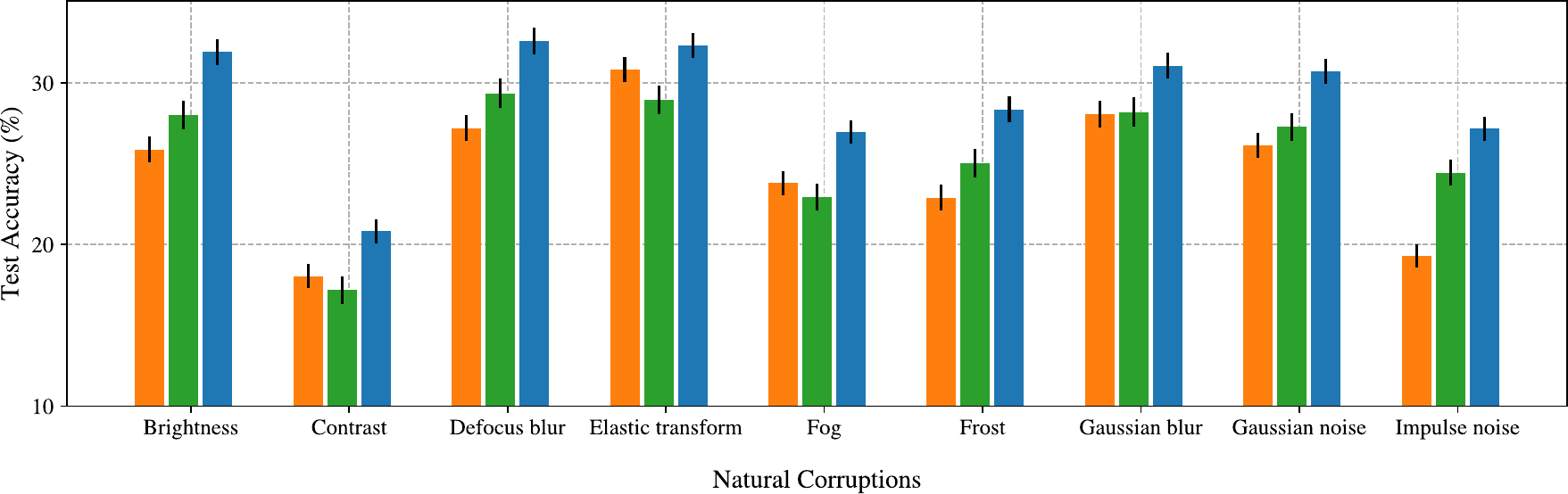}
    \caption{Robustness to natural corruptions.}
    \label{fig:corruption}
\end{figure*}

\paragraph{Robustness to natural corruptions} In the real world, autonomous cars must operate effectively in various environments, including diverse weather conditions. Therefore, it is essential for FCIL algorithms to be robust to naturally corrupted data distributions. We then evaluate our method and the recent FCIL methods on the \textbf{CIFAR-100-C} dataset. This dataset includes 18 augmentations of the original CIFAR-100, inspired by CIFAR-10-C~\cite{hendrycks2019benchmarking}. Models are trained using standard CIFAR-100 with the same setting in Section~\ref{sec:implementation} and tested on CIFAR-100-C. Figure~\ref{fig:corruption} shows robustness to 09 different corruptions averaged over three different runs, the results of the rest augmentations are shown in the Appendix~\ref{ap:corruption}. Specifically, our approach achieves higher test accuracy on various corruptions, with an average improvement of 5\% over MFCL and 8\% over TARGET. Evidently, our method offers noticeable advantages in robustness against natural corruption.

\begin{figure*}[t]
    \centering
    \includegraphics[scale=0.65]{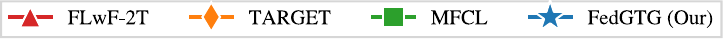}\\
    \vspace{0.2cm}
     \begin{subfigure}[b]{0.30\textwidth}
         \centering
         \includegraphics[scale=0.58]{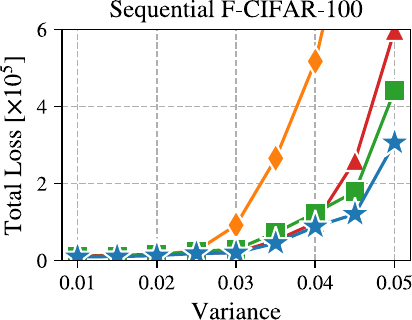}\\
         \caption{Perturbation [$\downarrow$]}
         \label{fig:minima:cifar100}
     \end{subfigure}
     \hfill
     \begin{subfigure}[b]{0.30\textwidth}
         \centering
         \includegraphics[scale=0.58]{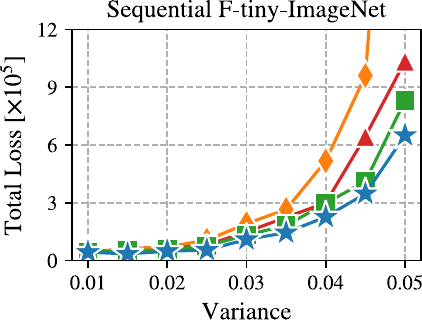}\\
         \caption{Perturbation [$\downarrow$]}
         \label{fig:minima:tinyimg}
     \end{subfigure}
     \hfill
     \begin{subfigure}[b]{0.30\textwidth}
         \centering
         \includegraphics[scale=0.58]{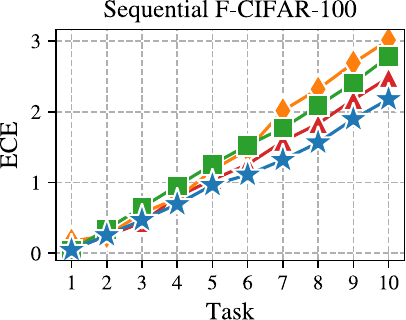}\\
         \caption{Calibration Error [$\downarrow$]}
         \label{fig:calibration:cifar100}
     \end{subfigure}
     \hfill
     \begin{subfigure}[b]{0.30\textwidth}
         \centering
         \vspace{0.2cm}
         \includegraphics[scale=0.58]{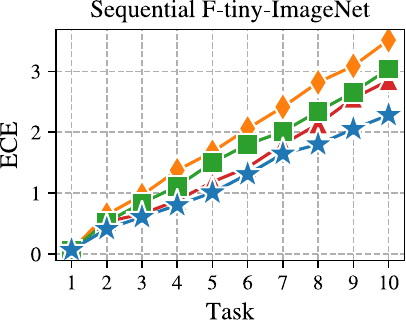}\\
         \caption{Calibration Error [$\downarrow$]}
         \label{fig:calibration:tinyimg}
     \end{subfigure}
     \hfill
     \begin{subfigure}[b]{0.30\textwidth}
         \centering
         \vspace{0.2cm}
         \includegraphics[scale=0.58]{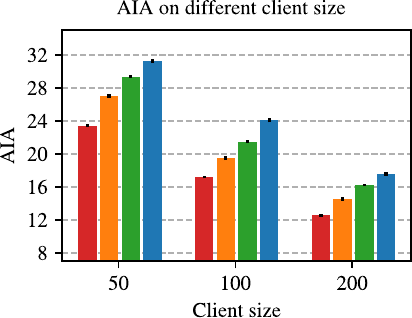}\\
         \caption{Average Accuracy [$\uparrow$]}
         \label{fig:size:aia}
     \end{subfigure}
     \hfill
     \begin{subfigure}[b]{0.30\textwidth}
         \centering
         \vspace{0.2cm}
         \includegraphics[scale=0.58]{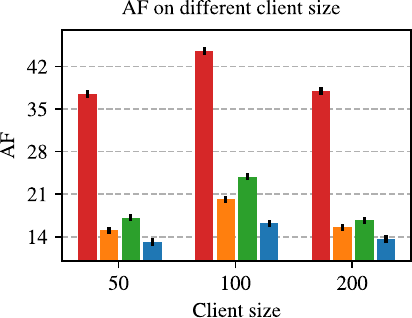}\\
         \caption{Average Forgetting [$\downarrow$]}
         \label{fig:size:af}
     \end{subfigure}
        \caption{Results for the model analysis. [$\uparrow$] higher is better, [$\downarrow$] lower is better \textit{(best seen in color)}.} 
        \label{fig:analysis}
\end{figure*}

\paragraph{Converging to flatter minima} Extensive CL algorithms~\cite{bhat2022task, wang2023metamix, park2024layer} explore how well their methods generalize by examining their ability to converge to flat minima. If a model obtains this quality, the loss function values $\mathcal{L}_{\text{CE}}$ increase only slightly, suggesting stable model predictions and demonstrating good train-test generalization. In this part, we compare the flatness of the training minima of FLwF-2T, TARGET, and MFCL with our approach. As done in~\cite{zhang2019your}, we consider the model at the end of training and add independent Gaussian noise with growing variance to each parameter. This allows us to evaluate its effect on the average loss $\sum_{t=1}^n\mathcal{L}^{\left(T_t\right)}_{CE}$ across all training examples. As shown in Figures~\ref{fig:minima:cifar100} and~\ref{fig:minima:tinyimg}, MFCL, especially FLwF-2T and TARGET, reveal higher sensitivity to perturbations than FedGTG. This result concludes that FedGTG can achieve better generalization compared to previous methods.

\paragraph{Converging to a more calibrated network} Calibration measures how well a learner’s prediction confidence matches its accuracy, with ideal outcomes reflecting true probabilities of correctness. In real-world applications, including weather forecasting~\cite{brocker2009reliability} and econometric analysis~\cite{gneiting2007probabilistic}, the calibrating ability of an FCIL algorithm should be investigated. Figures~\ref{fig:calibration:cifar100} and~\ref{fig:calibration:tinyimg} show the value of the Expected Calibration Error (ECE)~\cite{naeini2015obtaining} across various FCIL methods after completing each task. It can be seen that FedGTG achieves a lower ECE than the others. This indicates that models trained with FedGTG are less over-confident and, consequently, easier to interpret.

\paragraph{Robustness to different client sizes} We validate the effectiveness of FedGTG compared to other FCIL algorithms across different client sizes on the tiny-ImageNet dataset. We run experiments by varying the number of total clients (maintaining a consistent participation rate of $0.1$ per round), ranging from 50 to 200, and compare the results. Figures~\ref{fig:size:aia} and~\ref{fig:size:af} demonstrate that our method still outperforms other approaches, achieving an accuracy 4\% higher and a forgetting rate 6\% lower compared to the next best method, MFCL.

\subsection{Limitations}
\label{sec:limitation}

In our work, clients require generative models, the global model's weight from the most previous task, and the current global model, which raises storage burdens between the server and clients. However, there are more benefits than storing actual data. First, the memory required for the generative models does not depend on the class size: as the number of classes increases, clients may need to either delete some stored examples to make space for new ones or expand their memory capacity. In contrast, the size of the generative models remains constant. Moreover, clients can remove the generative models when inactive during a specific round and retrieve them later when rejoining, whereas removing data samples leads to a permanent loss of information.



\section{Conclusion}

This study introduces an FCIL framework that tackles the constraints of limited resources and privacy concerns. We utilize data and feature generative models that have been trained by the server, eliminating the requirement for costly on-device memory for clients. Our experiments provide evidence that our strategy is successful in reducing catastrophic forgetting and surpasses the current state-of-the-art methods. This paper also analyzes the robustness of FCIL algorithms on natural images, as well as testing the qualities of converging to flat minima and calibrated networks.


\newpage
{
    \small
    \bibliographystyle{unsrtnat}
    \bibliography{main}
}

\newpage
\appendix

\section{FedGTG algorithm}
Recall that there are $n$ tasks $T_1$, $T_2$, $\ldots$, $T_n$. At task $T_1$, the system is trained using the conventional FedAVG algorithm for aggregating the weight from the clients in $R$ communication rounds. At the end of every task, the server trains a data generator and a feature generator without using any information from the clients. From task $T_2$, these two generators are sent to the clients, which combine with modified objectives to both retain knowledge and learn new tasks well. We formalize our approach in Algorithm \ref{alg:fedgtg} in detail. The code is available at \url{https://github.com/lucaznguyen/FedGTG}.

\begin{algorithm*}[htbp]
\caption{Federated Global Twin Generator}
\label{alg:fedgtg}
\begin{algorithmic}[1]
\State \textbf{Input:}
\State $n$ tasks with $n$ datasets $\left\{\mathcal{D}_1, \mathcal{D}_2, \ldots, \mathcal{D}_n\right\}$.
\State $c$ clients with $c$ local models $\theta$, $R$ communication rounds.
\State A global model $\theta_G$, a data generator $G_D$ and a feature generator $G_F$.
\State \textbf{Procedure:}
\For{$t=1 \textbf{ to } T$}
    \For{$r=1 \textbf{ to } R$} \Comment Each task is learned on several communication rounds
        \State Select $k$ clients for training.
        \If {$r>1$ or $t>1$}
            \State The server sends the global model's weight to selected clients.
            \If {$t>1$}
                \State The server sends the two generators, the global model's weight from the previous task and the Empirical Feature Matrix to selected clients.
            \EndIf
        \EndIf
        \If{$t=1$}
            \State Train local models $\theta^{\left(t, r\right)}_j$ conventionally. \Comment{$1 \leq j \leq k$}
        \Else
            \State 
            \State Train local models $\theta^{\left(t, r\right)}_j$ using Algorithm~\ref{alg:client}.
        \EndIf
        \State Aggregate local model updates to the server.
    \EndFor
    \State Train the data generator and the feature generator.
    \State Calculate Empirical Feature Matrix $E^t$ using synthetic features.
\EndFor
\end{algorithmic}
\end{algorithm*}

\begin{algorithm*}[htbp]
\caption{Client-side: Continual Learning}
\label{alg:client}
\begin{algorithmic}[1]
\State \textbf{Input:}
\State Task $T_t$, $t\geq2$ with the dataset $\mathcal{D}_t$ in round $r$ has $B$ batches.
\State The global model $\theta_G^{\left(t, r\right)}$, a data generator $G_D^{t-1}$, a feature generator $G_F^{t-1}$.
\State The freezed global weight $\theta_G^{\left(t-1\right)}$ and the Empirical Feature Matrix $E^{t-1}$.
\State \textbf{Procedure:}
\State Calculate the Current Cross-Entropy Loss $\mathcal{L}_{\text{CE}}$ using $\mathcal{D}_t$ and Equation~\ref{loss:ce}.
\State Generate synthetic data $\mathcal{D}_S$ and synthetic features $\mathcal{F}_S$ having $B$ batches each.
\State Calculate the Fine-tunig Loss $\mathcal{L}_{\text{FT}}$ using $\mathcal{D}_t$, $\mathcal{D}_S$ and Equation~\ref{loss:ft}.
\State Calculate the Synthetic Logits Distillation Loss $\mathcal{L}_{\text{logits}}$ using $\mathcal{F}_S$ and Equation~\ref{loss:logit}.
\State Calculate the EFM Loss $\mathcal{L}_{\text{EFM}}$ using $\mathcal{F}_S$ and Equation~\ref{loss:efm}.
\State Optimize Equation~\ref{loss:optimize}.
\end{algorithmic}
\end{algorithm*}

\section{Experimental Setup}
\label{ap:setup}

In this section, we detail the settings used in our experiments, including datasets, FCIL algorithms, and experimental setups.

\paragraph{Datasets} We perform our experiments on three widely-used benchmark datasets, including the FCIL version of CIFAR-10~\cite{krizhevsky2009learning}, CIFAR-100~\cite{krizhevsky2009learning} and tiny-ImageNet~\cite{yao2015tiny}:

\begin{itemize}[leftmargin=*]
    \item {\bf Sequential F-CIFAR-10.}  The CIFAR-10 dataset~\cite{krizhevsky2009learning} consists of 60,000 $32\times32$ color images in 10 classes, with 6,000 images per class. There are 50,000 training images and 10,000 test images. We split the training set into five disjoint subsets corresponding to 5 tasks. 
    
    \item {\bf Sequential F-CIFAR-100.} Sequential F-CIFAR-100 is constructed by dividing the original CIFAR-100 dataset~\cite{krizhevsky2009learning}, which contains 50,000 images belonging to 100 classes, into ten disjoint subsets corresponding to 10 tasks. In this way, each task has 5,000 images from 10 distinct categories, and each class has 500 images. 
    
    \item {\bf Sequential F-tiny-ImageNet.} Tiny-ImageNet~\cite{yao2015tiny} is a subset of ImageNet, containing 100,000 images of 200 real objects. We follow settings in~\cite{babakniya2024data} to form the Sequential F-tiny-ImageNet. In particular, we split the original dataset into ten non-overlapping subsets. We consider each subset as a task whose images are labeled by 20 different classes, and each class has 500 samples.
\end{itemize}

\paragraph{FCIL Baselines} In addition to our FedGTG, we also include one regularization-based method, {\bf FLwF-2T}~\cite{usmanova2021distillation}, and two generative-based methods, {\bf TARGET}~\cite{zhang2023target} and {\bf MFCL}~\cite{babakniya2024data}. {\bf FLwF-2T} utilize knowledge distillation both on the server side and client side to ease the catastrophic forgetting issue. {\bf TARGET} utilizes a global model to transfer knowledge from past tasks to the current task while also training a generator to generate synthetic data, mimicking the overall data distribution across clients. {\bf MFCL} employs a generative model to create samples from previous distributions, which are then combined with training data to prevent catastrophic forgetting. Both of these algorithms ensure privacy by training the generative model on the server using data-free techniques after each task without client data retrieval. 

\paragraph{Implementation Details} Table~\ref{tab:setting} shows our settings and the hyper-parameter tuning for each dataset.

\section{Generative model setup}
\begin{figure}[htbp]
    \centering
    \begin{subfigure}[b]{1\textwidth}
    \centering
    \includegraphics[scale=0.5]{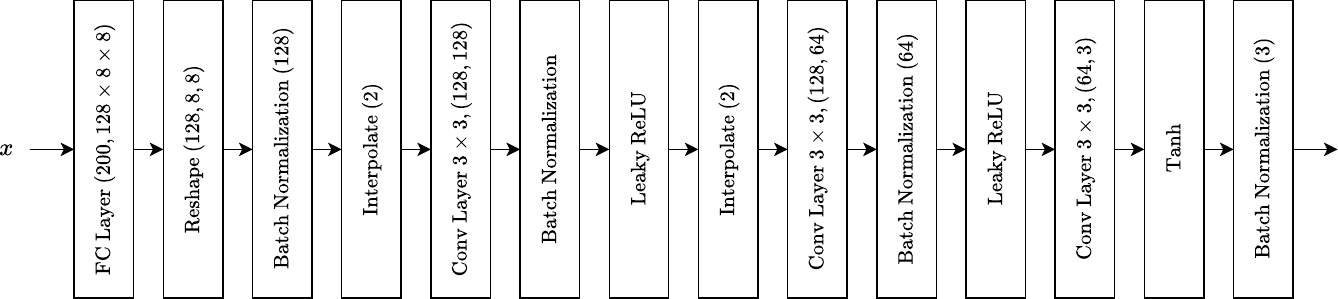}
    \caption{Sequential F-CIFAR-10 and Sequential F-CIFAR-100.}
    \end{subfigure}
    \begin{subfigure}[b]{1\textwidth}
    \centering
    \vspace{0.2cm}
    \includegraphics[scale=0.5]{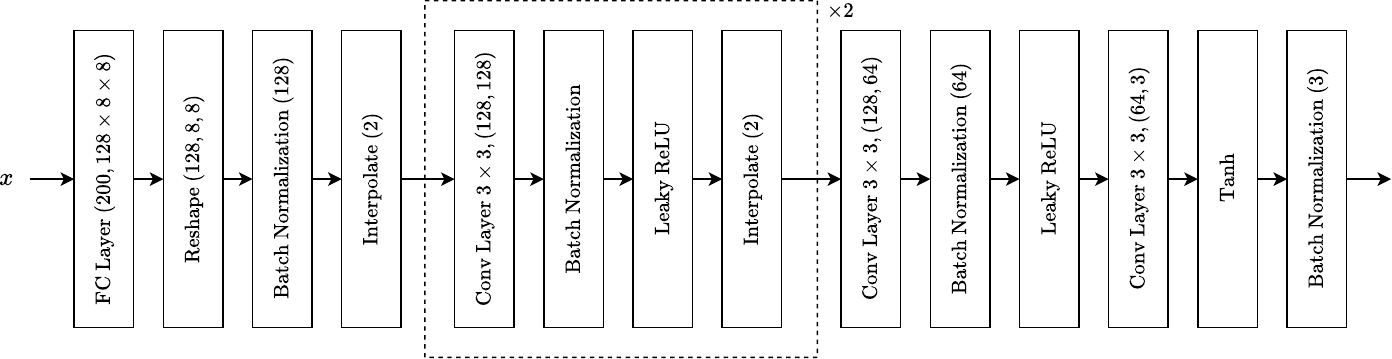}
    \caption{Sequential F-tiny-ImageNet.}
    \end{subfigure}
    \caption{Architecture of the data generative model across three datasets.}
    \label{fig:data:architecture}
\end{figure}

\paragraph{Data Generative Model Architecture}
Figure~\ref{fig:data:architecture} presents the architecture of the data generative models used for the Sequential F-CIFAR-10, Sequential F-CIFAR-100, and Sequential F-tiny-ImageNet dataset. In all experiments, the global model is based on the ResNet-18 backbone.

\begin{figure}[htbp]
    \centering
    \begin{subfigure}[b]{1\textwidth}
    \centering
    \includegraphics[scale=0.5]{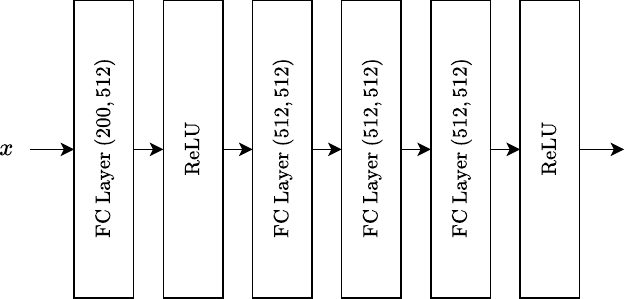}
    \caption{Sequential F-CIFAR-10 and Sequential F-CIFAR-100.}
    \end{subfigure}
    \begin{subfigure}[b]{1\textwidth}
    \centering
    \vspace{0.2cm}
    \includegraphics[scale=0.5]{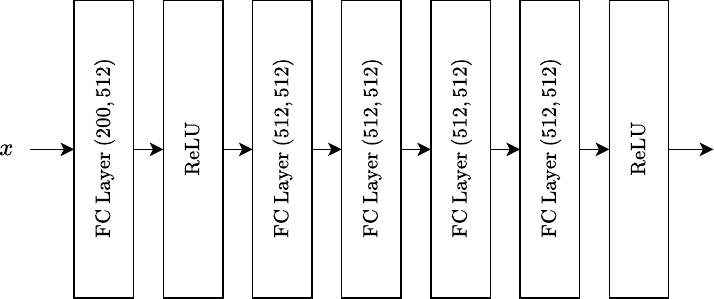}
    \caption{Sequential F-tiny-ImageNet.}
    \end{subfigure}
    \caption{Architecture of the data generative model across three datasets.}
    \label{fig:feature:architecture}
\end{figure}

\paragraph{Feature Generative Model Architecture} The architecure of the feature generative models is illustrated in Figure~\ref{fig:feature:architecture}, which employed for the Sequential F-CIFAR-10, Sequential F-CIFAR-100, and Sequential F-tiny-ImageNet datasets. As the outputs are feature vectors, only fully connected layers are needed.

\paragraph{Information generation}
To create synthetic data, clients sample i.i.d. noise, which is used to determine the classes through the application of the \texttt{argmax} function to the first $q$ elements, where $q$ represents the total number of classes observed. Since the noise is sampled i.i.d., each class has an equal probability of $\frac{1}{q}$ for sample generation.

\begin{table}[t]
\centering
\begin{tabular}{ccccc}
\hline
\textbf{Setting}                         & \textbf{Dataset}           & \textbf{CIFAR-10} & \textbf{CIFAR-100} & \textbf{tiny-ImageNet} \\ \hline
\multirow{10}{*}{Experimental setup}     & Image size                 & $32\times 32$     & $32\times 32$      & $64\times 64$          \\
                                         & Task number                & 5                 & 10                 & 10                     \\
                                         & Classes per task           & 2                 & 10                 & 20                     \\
                                         & Samples per task           & 5000              & 500                & 500                    \\
                                         & Learning rate              & 0.1               & 0.1                & 0.1                    \\
                                         & Weight decay               & 0.1               & 0.1                & 0.1                    \\
                                         & Batch size                 & 32                & 32                 & 32                     \\
                                         & Synthetic batch size       & 32               & 32                & 128                    \\
                                         & Communication round        & 100               & 100                & 100                    \\
                                         & Local epoch                & 10                & 10                 & 10                     \\ \hline
\multirow{10}{*}{Hyper-parameter tuning} & $\lambda_{\text{IE}}$     & 1.0               & 1.0                & 1.0                    \\
                                         & $\lambda_{\text{batch}}$   & 1.0               & 1.0                & 1.0                    \\
                                         & $\lambda_{\text{smooth}}$  & 1.0               & 1.0                & 1.0                    \\
                                         & $\lambda_{\text{FIE}}$    & 1.0               & 1.0                & 1.0                    \\
                                         & $\lambda_{\text{current}}$ & 1.5               & 1.5                & 2.0                    \\
                                         & $\lambda_{\text{FT}}$      & 1.0               & 1.0                & 1.0                    \\
                                         & $\lambda_{\text{logits}}$  & 0.1               & 0.1                & 0.05                   \\
                                         & $\lambda_{\text{EFM}}$     & 0.005             & 0.005              & 0.005                  \\
                                         & $\lambda_{\text{E}}$       & 10.0              & 10.0               & 10.0                   \\
                                         & $\eta$                     & 0.1               & 0.1                & 0.1                    \\ \hline
\end{tabular}
\vspace{0.3cm}
\caption{Detail settings across three datasets.}
\label{tab:setting}
\end{table}

\section{Additional results}
\label{ap:corruption}
In this section, we show additional results about the robustness of testing on natural images across our method and other FCIL methods. Figure~\ref{fig:corruption} shows the last 09 augmentations of the CIFAR-100 dataset averaged over three different runs. Our approach still outperforms MFCL and TARGET in terms of test accuracy.

\begin{figure*}[htbp]
    \centering
    \includegraphics[scale=0.65]{figure/legend-corruption.pdf}\\
    \vspace{0.2cm}
    \includegraphics[scale=0.7]{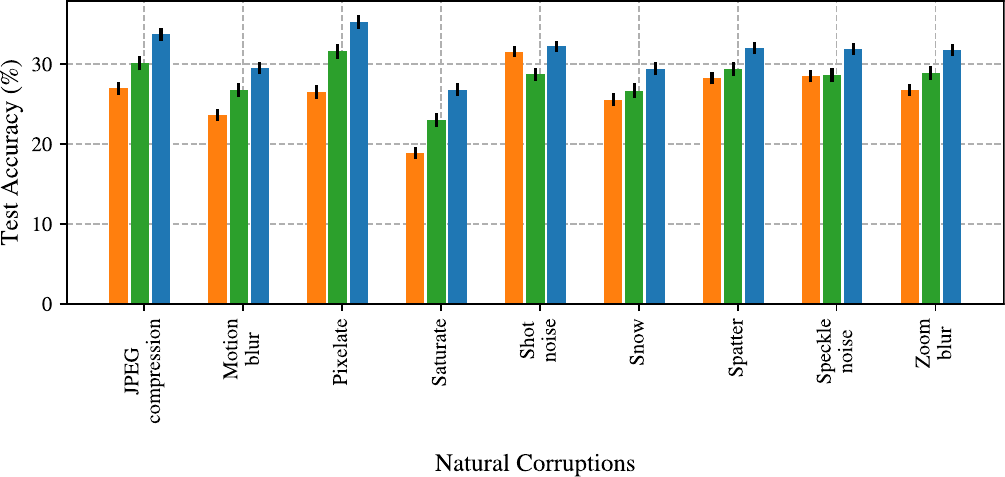}
    \caption{Robustness to natural corruptions.}
    \label{fig:corruption_last}
\end{figure*}

\section{Ablation Study}

\begin{table}[h]
\centering
\resizebox{\textwidth}{!}{
\begin{tabular}{ccccccccccccc}
\hline
\textbf{w/o Loss}             & $\mathbf{\mathcal{A}^1}$ & $\mathbf{\mathcal{A}^2}$ & $\mathbf{\mathcal{A}^3}$ & $\mathbf{\mathcal{A}^4}$ & $\mathbf{\mathcal{A}^5}$ & $\mathbf{\mathcal{A}^6}$ & $\mathbf{\mathcal{A}^7}$ & $\mathbf{\mathcal{A}^8}$ & $\mathbf{\mathcal{A}^9}$ & $\mathbf{\mathcal{A}^{10}}$ & $\mathbf{\mathcal{A}}$ & $\mathbf{\mathcal{F}}$ \\ \hline
$\mathcal{L}_{\text{IE}}$    & $72.40$                  & $56.60$                  & $46.96$                  & $38.22$                  & $33.18$                  & $30.73$                  & $28.50$                  & $25.05$                  & $23.98$                  & $22.37$                     & $33.95$                & $37.31$                \\
$\mathcal{L}_{\text{batch}}$  & $72.40$                  & $55.62$                  & $48.67$                  & $41.25$                  & $33.48$                  & $30.08$                  & $27.10$                  & $22.98$                  & $22.48$                  & $21.41$                     & $33.67$                & $41.42$                \\
$\mathcal{L}_{\text{smooth}}$ & $72.40$                  & $57.15$                  & $49.10$                  & $41.85$                  & $39.44$                  & $37.58$                  & $34.67$                  & $31.15$                  & $30.50$                  & $29.20$                     & $38.96$                & $29.36$                \\
$\mathcal{L}_{\text{FIE}}$   & $72.40$                  & $56.35$                  & $47.96$                  & $37.22$                  & $34.18$                  & $32.61$                  & $29.82$                  & $26.17$                  & $24.98$                  & $23.03$                     & $34.70$                & $33.42$                \\
$\mathcal{L}_{\text{FT}}$     & $72.40$                  & $41.80$                  & $34.67$                  & $29.25$                  & $22.68$                  & $17.60$                  & $15.17$                  & $13.06$                  & $12.59$                  & $12.09$                     & $22.10$                & $11.66$                \\
$\mathcal{L}_{\text{logits}}$ & $72.40$                  & $57.15$                  & $49.77$                  & $43.05$                  & $39.36$                  & $37.72$                  & $35.67$                  & $33.19$                  & $32.73$                  & $31.62$                     & $40.03$                & $25.35$                \\
$\mathcal{L}_{\text{EFM}}$    & $72.40$                  & $56.30$                  & $48.77$                  & $42.53$                  & $39.56$                  & $37.28$                  & $35.37$                  & $31.91$                  & $31.66$                  & $29.92$                     & $39.26$                & $27.38$                \\ \hline
FedGTG (ours)                 & $72.40$                  & $57.95$                  & $54.85$                  & $50.20$                  & $47.45$                  & $44.95$                  & $41.00$                  & $39.01$                  & $37.92$                  & $36.46$                     & $46.42$                & $18.66$                \\ \hline
\end{tabular}
}
\vspace{0.3cm}
\caption{Ablation study for FedGTG on Sequential F-CIFAR-100.}
\label{tab:ablation}
\end{table}

In this section, we highlight the significance of each loss within our proposed framework, analyzing both server and client contributions by sequentially removing components to observe their effects. Table~\ref{tab:ablation} shows our results, where each row corresponds to the removal of a specific loss component, and the columns display the corresponding Average Accuracy ($\mathcal{A}^t$), for $1\leq t\leq 10$, Average Incremental Accuracy ($\mathcal{A}$), and Average Forgetting ($\mathcal{F}$) from our proposed method. Specifically, we can see that the performance of the model is influenced by generative models, as poorly trained ones result in low AIA and high AF compared to others. Nevertheless, the Fine-tuning Loss has the lowest AF among all just because it did not learn tasks well (lowest AIA). The final two rows illustrate how the feature-constraining losses ($\mathcal{L}_{\text{logits}}$ and $\mathcal{L}_{\text{EFM}}$) impact the performance of the global model, where the decrease in accuracy demonstrates the importance of these two losses.

\end{document}